%% file: 000_main.tex
\documentclass[journal,twoside,web]{ieeecolor}
\usepackage{generic}
\usepackage{cite}
\usepackage{amsmath,amssymb,amsfonts}
\usepackage{graphicx}
\usepackage{hyperref}
\hypersetup{hidelinks=true}
\usepackage{textcomp}
\usepackage{makecell}
\usepackage{booktabs}
\usepackage{tabularx}
\usepackage{bm}
\usepackage{multirow}
\usepackage{algorithm}%
\usepackage{algpseudocode}%
\usepackage{subfigure}

\newcommand{\citep}[1]{\cite{#1}}
\usepackage{xurl}

\usepackage{xspace}
\newcommand{\name}{%
{\textsc{PosNegDM}}%
\xspace
}
\newcommand{\tname}{%
{\textsc{DualSight}}%
\xspace
}

\def\BibTeX{{\rm B\kern-.05em{\sc i\kern-.025em b}\kern-.08em
    T\kern-.1667em\lower.7ex\hbox{E}\kern-.125emX}}
\begin{document}

\title{Reinforced Sequential Decision-Making for Sepsis Treatment: The \name Framework with Mortality Classifier and Transformer}

\author{Dipesh Tamboli, Jiayu Chen, Kiran Pranesh Jotheeswaran, Denny Yu, \\and Vaneet Aggarwal \thanks{The authors are with Purdue University, West Lafayette, IN 47906, USA, email: \{dtamboli, chen3686, kjothees, dennyyu, vaneet\}@purdue.edu. This work was supported in part by the National Science Foundation under Grant No. 2129097. \\This is the final author version of the paper accepted to IEEE Journal of Biomedical and Health Informatics, Mar 2024. \\© 2024 IEEE. Personal use of this material is permitted. Permission from IEEE must be obtained for all other uses, in any current or future media, including reprinting/republishing this material for advertising or promotional purposes, creating new collective works, for resale or redistribution to servers or lists, or reuse of any copyrighted component of this work in other works.
}}

% \author{First A. Author, \IEEEmembership{Fellow, IEEE}, Second B. Author, and Third C. Author Jr., \IEEEmembership{Member, IEEE}
% \thanks{This paragraph of the first footnote will contain the date on 
% which you submitted your paper for review. It will also contain support 
% information, including sponsor and financial support acknowledgment. For 
% example, ``This work was supported in part by the U.S. Department of 
% Commerce under Grant 123456.'' }
% \thanks{The next few paragraphs should contain 
% the authors' current affiliations, including current address and e-mail. For 
% example, First A. Author is with the National Institute of Standards and 
% Technology, Boulder, CO 80305 USA (e-mail: author@boulder.nist.gov). }
% \thanks{Second B. Author Jr. was with Rice University, Houston, TX 77005 USA. He is 
% now with the Department of Physics, Colorado State University, Fort Collins, 
% CO 80523 USA (e-mail: author@lamar.colostate.edu).}
% \thanks{Third C. Author is with 
% the Electrical Engineering Department, University of Colorado, Boulder, CO 
% 80309 USA, on leave from the National Research Institute for Metals, 
% Tsukuba, Japan (e-mail: author@nrim.go.jp).}}

\maketitle

\input{001_abstract}

\input{01_intro}

\input{02_related}

\input{03_approach}
\input{04_data}

\input{05_evaluation}

\input{06_conclusion}

\section*{References}

\bibliographystyle{plain}
\bibliography{sn-bibliography}

\clearpage
\input{071_appendix}

\end{document}

%% file: 001_abstract.tex
\begin{abstract}
Sepsis, a life-threatening condition triggered by the body's exaggerated response to infection, demands urgent intervention to prevent severe complications. Existing machine learning methods for managing sepsis struggle in offline scenarios, exhibiting suboptimal performance with survival rates below 50\%. This paper introduces the \name ---
``Reinforcement Learning with Positive and Negative Demonstrations for Sequential Decision-Making" framework utilizing an innovative transformer-based model and a feedback reinforcer to replicate expert actions while considering individual patient characteristics. A mortality classifier with 96.7\% accuracy guides treatment decisions towards positive outcomes. The \name framework significantly improves patient survival, saving 97.39\% of patients, outperforming established machine learning algorithms (Decision Transformer and Behavioral Cloning) with survival rates of 33.4\% and 43.5\%, respectively. Additionally, 
ablation studies underscore the critical role of the transformer-based decision maker and the integration of a mortality classifier in enhancing overall survival rates. In summary, our proposed approach presents a promising avenue for enhancing sepsis treatment outcomes, contributing to improved patient care and reduced healthcare costs.

\end{abstract}

\begin{IEEEkeywords}
Machine Learning, Transformer, Sepsis Treatment, Healthcare
\end{IEEEkeywords}

%% file: 01_intro.tex
\section{Introduction}
\label{sec:introduction}

Sepsis is a life-threatening medical condition characterized by acute organ dysfunction. According to \cite{CDC2023}, in a typical year, at least 1.7 million adults in America develop sepsis and at least 350,000 adults who develop sepsis die during their hospitalization or are discharged to hospice. Further, 1 in 3 people who dies in a hospital had sepsis during that hospitalization. Early intervention and appropriate treatment are critical for reducing mortality rates. while the current guidelines for sepsis treatment, such as the Surviving Sepsis Campaign \cite{b19}, have played a significant role in improving patient outcomes, they also come with their own set of challenges \cite{Fleischmann-Struzek2020-lp}. The primary disadvantage of these standards lies in their one-size-fits-all approach, which may not cater to the unique characteristics and varying responses of individual patients \cite{rhee2022heterogeneity}. Furthermore, current methods may not fully take into account the rapidly evolving nature of sepsis, leading to delayed or inadequate treatment modifications \cite{ranjit2021challenges}. In addressing these challenges, our paper focuses on a learning-based methodology that grounds a continuous treatment strategy in the ongoing condition of the patient. Employing a Reinforcement Learning (RL) approach \cite{b23, b24}, we leverage its capacity to adapt and optimize treatment plans over time. This affords a dynamic and personalized approach, closely aligned with the evolving needs of the individual.

RL has demonstrated success in addressing medical issues, such as diabetes \citep{b18,b20} and mechanical ventilation in Intensive Care Units (ICUs) \citep{b21}. RL can effectively model physicians' decision-making processes by leveraging historical patient data and past medical decisions. This allows it to guide real-time treatment decision-making, such as the optimization of drug dosages and the timing of administration \cite{b30}. Nevertheless, the application of online RL methods is constrained in practice due to the lack of a simulator providing the subsequent state based on the current patient state and treatment decision. This constraint extends to popular Imitation Learning (IL) techniques like GAIL \citep{ho2016generative} and AIRL \citep{fulearning}, which necessitate a similar simulator.
To overcome these challenges, we delve into the realm of Offline RL (ORL) \cite{b22}. Unlike traditional RL, which learns from real-time interactions, offline RL leverages pre-collected datasets, making it particularly relevant for sepsis treatment. In our context, the dataset comprises trajectories representing sequences of states, actions, and outcomes observed in past Sepsis cases. This shift to offline RL allows us to learn from both positive and negative trajectories, providing a more comprehensive understanding of effective treatment strategies. The utilization of offline RL is a key aspect of our proposed \name framework, enhancing its ability to derive impactful insights from historical patient data.

Specifically, we propose the \name framework, a novel approach to assist physicians in identifying effective treatment strategies for sepsis by selectively learning from past expert treatment decisions. 
\name stands for ``Reinforcement Learning with Positive and Negative Demonstrations for Sequential Decision-Making." It is an offline RL-based model that leverages both positive and negative trajectories to improve sepsis treatment.
The proposed \name architecture is built upon a Transformer model \citep{b12} and includes a feedback reinforcer for performance improvement. With \name, the agent can learn from sequences of decisions leading to positive outcomes while concurrently deviating sequences that result in negative consequences, which is accomplished by promoting treatment decisions that steer towards positive outcomes, such as decreased mortality rates.
The key components of the proposed architecture are (i) a mortality classifier, (ii) a Transformer-based model that predicts the next state and action based on the historical patient states and treatments, and (iii) an efficient loss function that includes both next-state and next-action prediction loss as well as a reinforced feedback loss based on the mortality classifier. 

As shown in Figure \ref{fig:intro}, 
the \name framework consists of two main components: a mortality classifier (also works as a feedback reinforcer), and a transformer-based decision maker.
The mortality classifier is responsible for predicting the probability of mortality for a given patient based on their clinical data. The transformer-based decision maker takes the patient's clinical data as input and generates a sequence of treatment decisions along with the next states. 
The feedback reinforcer (trained and frozen Mortality Classifier) takes the next states, evaluates the effectiveness of the treatment decisions, and provides feedback ($L_{survival}$ loss, Eq. \ref{eq:adversarial}) to the decision maker to improve future decisions. Once the Decision Maker is ready, it generates actions that lead to the live states and save patients.\\

This is the first offline RL architecture that aims to reduce mortality by explicitly learning to avoid negative trajectories, making it a promising approach to improve Sepsis treatment. This paper takes a significant step forward by quantifying mortality rates and demonstrating that approaches achieving strong action prediction may not necessarily result in improved mortality outcomes.  The key contributions of the work can be summarized as:
\begin{enumerate}
    \item This paper introduces the \name framework, the first offline RL-based model that leverages both positive and negative trajectories to lower mortality rates in sepsis patients. Our innovative design incorporates a transformer-based architecture, a neural network-driven mortality classifier that acts as a feedback reinforcer, and a refined loss function that explicitly guides the transformer to circumvent negative trajectories.
	\item Leveraging a meticulously selected set of features, our proposed mortality classifier achieved a remarkable accuracy of 96.7\% in predicting mortality among sepsis patients.
	\item Our classification-feedback-reinforced transformer architecture is capable of emulating expert decision-making  and is shown to save 97.39\% patients, a significant improvement compared to the 33.4\% (DT) and 43.5\% (BC) survival rates. These results underscore the effectiveness of the \name framework. In real-world settings, our framework can serve as a valuable tool for physicians, providing sepsis treatment decisions as a reference based on the real-time state of patients.
\end{enumerate}

The article is structured to provide a comprehensive exploration of sepsis treatment and the application of \name. The introduction (Section 1) offers an overview of sepsis and delineates the treatment challenges associated with it, introducing the application of machine learning as a potential solution to enhance treatment outcomes. Section 2, the related work, reviews existing literature on machine learning in sepsis treatment, shedding light on the limitations of current approaches. Section 3, the methodology, intricately describes the \name framework, elucidating its components, such as the transformer-based model (\tname), the mortality classifier (feedback reinforcer). In Section 4, the experimental results showcase the performance of the \name framework and provide a comparative analysis with existing machine learning-based algorithms. Section 5, the discussion, delves into the implications of the results and outlines potential avenues for future research in this domain. Finally, Section 6, the conclusion, succinctly summarizes the key findings of the paper and emphasizes the significance of leveraging machine learning for improved sepsis treatment outcomes.

%% file: 02_related.tex
\section{Related Works}
\label{sec:related_works}

This section provides an overview of cutting-edge Machine Learning (ML) algorithms applied to healthcare decision-making, particularly focusing on Offline Reinforcement Learning and Imitation Learning. 

\subsection{Offline Reinforcement Learning}

The Markov Decision Process (MDP) can be represented as a tuple $(S, A, P, R)$ with states $s \in S$, actions $a \in A$, transition dynamics $P(s^{\prime}|s,a) \in [0, 1]$, and the reward function given by $r = R(s,a)$. A sequence of states, actions, and rewards would comprise a trajectory, which can be represented as $\tau = (s_1,a_1,r_1,\cdots,s_T,a_T,r_T)$. A reinforcement learning (RL) problem aims to learn a policy such that it maximizes the expected return $\mathbb{E}[\sum_{t=1}^T r_t]$. 
Offline RL \cite{b22} is proposed to train RL agents using pre-collected data rather than interacting with the environment in real time. This is in contrast to online RL \cite{b23, b24}, where the agent learns from the interactions with a simulator. Offline RL proves particularly beneficial when real-time data collection via a simulator is challenging or costly. It also presents a more practical solution, given the easier accessibility of trajectories.

In offline RL, an agent is trained on a dataset of previous interactions, often referred to as a replay buffer. This buffer contains tuples composed of the elements: state, action, reward, and next state. Such offline data can typically be gathered from people's routine practice for a given task. Using this information, the agent can learn a policy that correlates states with actions, and a value function that estimates the expected return of a certain state or state-action pair. The derived policy should ideally produce decisions that maximize future return from the current state. 

In recent literature on sepsis identification and treatment systems based on  offline RL, several approaches have been explored. These approaches aim to model the discrete Markov Decision Process (MDP) \cite{b30,raghu2017b,b4,b11}, and then use online RL approaches. However, the discrete feature space may lead to information loss, which affects the accuracy of online RL. Recent works \cite{raghu2018b,wang2022learning} have modeled MDP in continuous state, while the result relies on the model approximation. In this work, we aim to make decisions without learning the model. Further, these  works do not explicitly avoid the negative trajectories. On the contrary, \name selects actions from the given set to diverge from the negative trajectories, achieved by maximizing the divergence between the predicted and ground-truth trajectory.

\subsection{Imitation Learning and Behavioral Cloning}

Imitation Learning (IL) \cite{b25} is a method of training an agent to perform a task by observing and mimicking decision-making sequences from domain experts. IL can be employed to learn a broad spectrum of tasks, ranging from simpler ones such as line-following to more complex tasks like game-playing. 

Behavioral Cloning (BC) represents a specific IL method that trains an agent to mimic expert behavior by learning a policy identical to that of the expert. 
This employs supervised learning, treating states (\(x\)) as input data and actions (\(y\)) as target values. For continuous actions, the Mean Squared Error (MSE) loss (\(\text{E}((f(s)-a)^2)\)) is used, minimizing the difference between predicted (\(f(s)\)) and actual actions (\(a\)). In the case of discrete actions, a cross-entropy loss is employed, measuring the dissimilarity between predicted action probabilities and true action distributions. BC learns to replicate the expert's actions, facilitating the mapping from states to actions through supervised learning.
One of the earliest works on BC was proposed by \cite{b14} for autopilot applications. Since then, it has been broadly adopted in areas such as autonomous vehicles \citep{b15}, robotic manipulators \citep{b16}, and healthcare \citep{b17}. In this paper, we employ BC as one of the baseline methods for comparison. However, only decision trajectories leading to positive outcomes are used in the training process, as the agent cannot follow a negative trajectory wherein patients did not survive. Compared with BC, our proposed algorithm demonstrates superior data efficiency by explicitly incorporating negative samples from the dataset.

Negative trajectories can provide crucial insights into which actions should be avoided to prevent repetition of mistakes. Authors of \citep{b10} and \citep{b13} propose IL models that employ adversarial discriminators to convert negative trajectories into positive ones. However, these techniques necessitate access to simulators, much like the state-of-the-art (SOTA) IL algorithms GAIL \cite{ho2016generative} and AIRL \cite{fulearning}. In contrast, our proposed method does not depend on a simulator which can be expensive to construct and inaccessible in many application domains. Instead, it recovers a policy from offline data only, facilitated by specially designed objective functions. For instance, when dealing with negative trajectories, we select actions from the given set to diverge from the negative trajectories, achieved by maximizing the divergence between the predicted and ground-truth trajectory.

In our approach, we introduce a novel perspective in Sepsis treatment by explicitly leveraging negative trajectories, a strategy not explored in existing research. While conventional machine learning models, such as Decision Transformer (DT) \cite{b1} and Behavioral Cloning (BC) \cite{b14}, have demonstrated high action prediction accuracy, they lack the explicit consideration of negative outcomes. Our novel framework, named \name, explicitly addresses this gap by incorporating a Mortality Classifier that also acts as a  Feedback Reinforcer. The Mortality Classifier allows us to assess the potential negative outcome (mortality) of each action, enabling the reinforcement of the model to avoid actions that lead to unfavorable states. This explicit focus on negative trajectories distinguishes our work from previous studies and contributes to the development of a more robust and clinically relevant Sepsis treatment model. The decision to compare our model against DT and BC is motivated by their popularity and effectiveness in predicting actions; however, their omission of explicit consideration for negative trajectories positions our \name framework as an innovative and advantageous alternative in improving patient outcomes in Sepsis treatment.

%% file: 03_approach.tex
\section{Proposed Approach}
\label{sec:proposed_approach}

In this section, we first delineate the algorithm's components, which include the mortality classifier and the transformer-based dualsight network. Subsequently, we present the complete \name framework, with a particular focus on the novel training objectives.

\subsection{Mortality Classifier}
\label{subsec:mort_class}

We have devised a classifier capable of predicting a patient's mortality likelihood, discerning between probable survival or demise. The classifier takes the current patient state as input and produces a binary result signifying the survival likelihood. The architecture of our mortality classifier comprises five fully-connected layers of size 64, with the training hyperparameters outlined in Table \ref{cls_para}. A significant challenge encountered during the training process was the presence of a highly imbalanced dataset, with a significantly lower number of deceased patients compared to survivors. To rectify this, we employed Borderline SMOTE \cite{b26} to impute minority class data and upsampled the training set.

\begin{table}[htbp] 
\centering
\caption{Hyperparameters for training the mortality classifier}
\resizebox{.48\textwidth}{!}{
\begin{tabular}{|c||c|c|c|c|}
\hline
Parameter & Learning Rate & Weight Decay & Optimizer & Dropout \\
\hline
Value & 1e-3 & 1e-5 & Adam & 0.2 \\
\hline
\end{tabular}\label{cls_para}}
\end{table}

This classifier is trained to achieve high accuracy in predicting the outcome of a treatment trajectory. Empirically, it has demonstrated an impressive accuracy of 96.7\% on the test dataset. It is subsequently utilized to evaluate the quality of treatment decisions derived from various algorithms and to provide feedback for the training of the transformer-based decision maker (as represented in the loss function illustrated in Equation \eqref{eq:adversarial}).

\subsection{\tname Decision Maker}
\label{subsec:dualsight}

% \subsubsection{Actions, States and Rewards}
% \label{subsubsec:act_rew}

A transformer is particularly well-suited for sequential decision-making tasks, such as Sepsis treatment, due to its ability to capture dependencies and relationships between different states in a sequence effectively. The transformer-based model in the \name framework learns from historical data and expert actions by processing input sequences through modality-specific linear embeddings and a positional episodic timestep encoding. The tokens are then introduced into a GPT architecture, which anticipates output in an autoregressive fashion, employing a causal self-attention mask. This self-attention module calculates a weighted sum of the input states, with weights assigned based on the similarity among the states, allowing the model to capture dependencies and relationships between different states in the sequence effectively. The significance of this approach is that it enables the model to learn from both positive and negative trajectories, which is key to improving patient survival rates in sepsis treatment. Additionally, the feedback reinforcer in the \name framework uses the forecasted next states to ascertain the ultimate survival likelihood of patients, which is subsequently utilized as feedback to fortify the decision maker. This unique configuration significantly enhances the decision-maker's performance.

The \tname decision maker is trained to produce a sequence of expert-like treatment decisions based on the patient states. Training is done using offline data provided in the MIMIC-III dataset, detailed in Section \ref{data}. The actions in the RL setting represent medical interventions, including the administration of intravenous fluids and vasopressor drugs within a 4-hour window. These actions are defined within a $5 \times 5$ discrete space, where 0 denotes no drug administration, and the other four non-zero drug dosages are classified into four quartiles. This results in 25 possible discrete action combinations for the two treatments. Regarding the reward function, it assigns $+1$ for positive trajectories and $-1$ for negative ones, provided only at the final time step of the patient's trajectory. A trajectory is considered positive if the patient survives post-treatment, else it is deemed negative. The agent's ultimate goal is to maximize the cumulative reward throughout the entire treatment period.

\begin{figure}	\centerline{\includegraphics[scale=0.1]{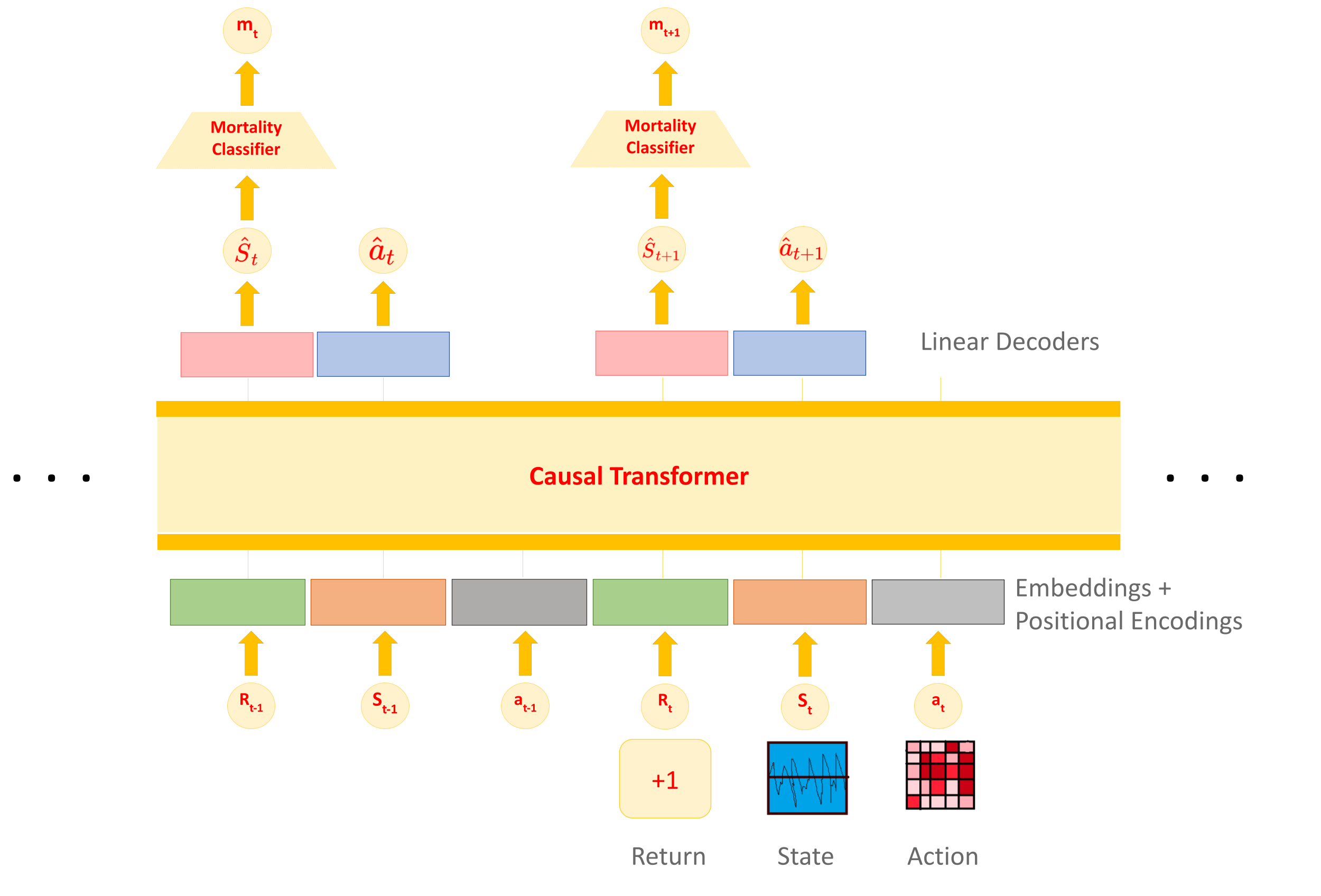}}
	\caption{The \tname decision maker takes in states, actions, and returns as input, which are first embedded into linear representations that are specific to each modality. The positional episodic timestep encoding is added to the input to help the model understand the order of events. The tokens are then fed into the GPT architecture, which uses a self-attention mechanism to predict actions and next states. The causal mask ensures that the model can only attend to previous tokens, preserving the causality of the system. The predicted states are subsequently input into the trained Mortality Classifier to assess whether the implemented action guides the patient towards a deceased state. The mortality prediction \(m_t\) is employed to influence the \tname, compelling it to choose actions aligned with the mortality classifier's prediction of an alive state. This process integrates mortality considerations into the decision-making mechanism, emphasizing the importance of actions that contribute to favorable patient outcomes.}
	\label{fig:intro}
\end{figure}

We utilize the network structure suggested in the Decision Transformer (DT) \cite{b1} for our decision maker. The Transformer model, renowned for its efficacy in sequence processing tasks, has been employed in leading Large Language Models like GPT \cite{b28} and BERT \cite{b27}. Leveraging a GPT-like architecture, DT has demonstrated exceptional performance, surpassing state-of-the-art offline RL baselines in numerous benchmarks.

As illustrated in Figure \ref{fig:intro}, our decision maker utilizes a sequence of past states, actions, and returns as input. The return signifies the aggregated future rewards from a given time step $t$ until the episode's termination, encapsulating the expected outcome over the decision enacted. This input sequence undergoes processing through modality-specific linear embeddings, supplemented with a positional episodic timestep encoding  to help the model understand the order of events. Subsequently, the tokens are introduced into a GPT architecture, which anticipates output in an autoregressive fashion, employing a causal self-attention mask. This self-attention module calculates a weighted sum of the input states, with weights assigned based on the similarity among the states, which allows the model to capture dependencies and relationships between different states in the sequence effectively. 

A crucial distinction between our architecture and that of the Decision Transformer lies in our decision maker's ability to anticipate not only the immediate action but also the ensuing state, thus offering dual insights. The forecasted next states are employed by the mortality classifier to ascertain the ultimate survival likelihood of patients, subsequently utilized as feedback to fortify the decision maker. This unique configuration significantly enhances our decision maker's performance.

\subsection{The Overall Framework: \name}
\label{subsec:overall_framework}

The key idea of \name architecture is to learn from positive trajectories and move away from negative ones. The \name framework offers insightful treatment recommendations, including the administration of intravenous fluids and vasopressor drugs. These suggestions guide clinicians by providing clear actions to consider and potential pitfalls to avoid, thereby bolstering the reliability and safety of their decision-making process. The \name architecture comprises two components: (a) an autoregressive decision maker that anticipates the current treatment decision (i.e., action) and subsequent patient states based on historical data, and (b) a mortality classifier that reinforces the decision maker if the final state produced by the preceding networks guides the patient towards a survival state. 
The Mortality Classifier is independently trained to distinguish between alive and deceased states. In the training of our decision-maker, the Mortality Classifier operates in inference (Feedback Reinforcer) mode rather than training mode to avoid interfering with \tname's learning. This enables the \tname to produce additional gradients derived from the alive/dead state Feedback Reinforcer's adversarial loss (Eq. \ref{eq:adversarial}). These supplementary gradients facilitate improved decision-maker training, enhancing its ability to predict actions that result in alive states and consequently reducing the mortality rate. The objectives for training the Decision Maker with the offline data and mortality classifier are shown as follows.

First, the decision maker is trained to predict the next state by minimizing the Mean Squared Error (MSE) between the predicted state $\hat{s}_t$ and the ground-truth state $s_t$:
\begin{equation} \label{eq:state_loss}
	L_{state}(\hat{s}_t, s_t) = L_{MSE}(\hat{s}_t, s_t) = \frac{1}{|S|} \sum_{i=1}^{|S|} (\hat{s}_{t,i} - s_{t,i})^2,
\end{equation}
where $|S|$ is the number of elements in the state vector. As outlined earlier, the forecast accuracy for subsequent states provides extra constraints in training the decision maker, when compared to the Decision Transformer. This design allows the decision maker to extract more relevant and beneficial features from the input, leading to a better equipped model to predict future treatment outcomes. Additionally, the anticipated final state will be fed into the mortality classifier to calculate patient survival rates, which further augments the training feedback.

Next, in order to prevent negative outcomes, the final patient state guided by the sequence of treatments should exhibit a relatively high survival rate. Utilizing the pretrained mortality classifier, we can incorporate another objective that acts as a constraint on survival outcomes, making treatment decisions safer and more reliable. Specifically, we introduce the loss term as follows:
\begin{align} \label{eq:adversarial}
&L_{survival}(MC,\tname, \tau)\nonumber\\ &= \mathbb{E}_{\tau \sim P_{data}(\cdot)}[\log(1 - MC(\tname(\tau)))]
\end{align}
Here, $\tau$ denotes the decision sequence sampled from an offline dataset with the distribution $P_{data}(\cdot)$, $\tname(\tau)$ represents the predicted final patient state  following a series of treatments provided by the decision maker, $MC$ represents the mortality classifier that gives out the survival rate. The training of the decision maker is geared towards enhancing the patient's survival rate, which in turn minimizes the survival loss $L_{survival}$.

Last, we minimize the action prediction error in the positive trajectories and maximize it in the negative ones. By doing so, we encourage our model to formulate decision rules that not only mimic the expert's actions but also learn to evade actions tied to negative outcomes. This results in a more robust understanding of action-avoidance related to negative trajectories. The loss associated with action prediction is formulated as:
\begin{equation} \label{eq:action_loss}
	L_{action}(\hat{a}_t^+, a_t^+,\hat{a}_t^-, a_t^-) = L_{CE}(\hat{a}_t^+, a_t^+) + \eta L_{CE}(\hat{a}_t^-, a_t^-)
\end{equation}
In this equation, $a^+$ corresponds to the actions involved in positive trajectories, whereas $a^-$ represents the actions derived from negative trajectories. $L_{CE}$ represents the cross-entropy loss, and $\eta$ is a weight factor that assists us in maintaining a balance between adhering to expert trajectories and circumventing negative outcomes.

Integrating the three objectives listed above, we can get an overall loss function as:
\begin{equation} \label{eq:loss_total}
	L_{total} = \alpha L_{action} + \beta L_{state} + \gamma L_{survival} 
\end{equation}
where the parameters $\alpha, \beta, \gamma$ control the weight of each term in the overall loss.
Rather than merely duplicating treatment decisions (i.e., actions), we propose a three-fold improvement: (1) Empower the agent to predict subsequent states, enriching its understanding of the input information. (2) Explicitly improve the survival rate as assessed by the mortality classifier. (3) Distinctly address action predictions for positive and negative trajectories. As a result, our algorithm can make better use of the treatment sequences, which includes a mixture of positive and negative trajectories, to achieve a significantly higher survival rate.

Algorithm \ref{alg:posnegdm} outlines the training procedure for the POSNEGDM framework, consisting of a Mortality Classifier (MC) and a Decision Maker (DM). The Mortality Classifier is initially trained on the mortality ground truth using cross-entropy loss. The trained Mortality Classifier, denoted as $w_{T_{MC}}$, is then frozen and serves as the Feedback Reinforcer. Subsequently, the Decision Maker is trained using a combination of losses, including survival loss ($\mathcal{L}_{survival}$) adversarially enforced by the Mortality Classifier. The DM's weights, denoted as $w_{T_{DM}}$, are updated iteratively based on the overall loss calculated from the mortality prediction, state prediction, and action prediction. This process aims to enhance the Decision Maker's ability to predict actions leading to patient survival by leveraging the adversarial feedback from the Mortality Classifier.

\input{03_z_algorithm}

%% file: 03_z_algorithm.tex
\begin{algorithm}[ht]
  \caption{\name}
  \label{alg:posnegdm}

\begin{algorithmic}[1]
\State $MC$: Mortality Classifier, $DM$: Decision Maker, $R_t$: Reward at timestep $t$ $\in \{-1, 1\}$, $a_t$: Action at timestep $t$, $S_t$: State at timestep $t$, $T_{MC}$: Total iterations for $MC$, $T_{DM}$: Total iterations for $MC$, $FR$: Feedback Reinforcer with frozen $MC$ weights, $M$: Mortality GroundTruth, $m$: Predicted Mortality, $CEL$: Cross-Entropy Loss, $\nabla g_w$: gradient.

\State {\bf Training Mortality Classifier}

\State Initialize $w_{MC}$ randomly
\For{$t = 1$ {\bfseries to} $T_{MC}$}
\State $S, M  \gets$ sample batch from $data_{train}$
\State $m \gets MC(S, w_{t_{MC}})$
\State $\mathcal{L} \gets CEL (m, M)$
\State Calculate $\nabla g_{w_{t_{MC}}}$ from $\mathcal{L}$ and update $w_{t_{MC}}$
\EndFor
\State $w_{T_{MC}}$ will be trained the Mortality Classifier
\State Frozen $\overline{w_{T_{MC}}}$ will act as the Feedback Reinforcer

\State {\bf Training \name}
\State Initialize $w_{DM}$ randomly

\For{$t = 1$ {\bfseries to} $T_{DM}$}
\State $d_{batch} \gets$ sample from $data_{train}$
  \State $S_t, a_t \gets d_{batch}$ 
  % \Comment{States and Actions from the batch}
  \State $S_{t+1}, a_{t+1} \gets DM(S_t, a_t, w_{t_{DM}})$ 
  % \Comment{Obtaining next action and state from the Decision Maker}
  \State $m_{t+1} \gets MC(S_{t+1}, \overline{w_{T_{MC}}})$
  \State $\mathcal{L}_{survival} \gets CEL(m_{t+1}, 1)$ (Forcing $m_{t+1}$ to be 1 
  \Statex \quad\  to adversarially train DM to take actions resulting in 
  \Statex \quad\ alive state) 
  % \Comment{}
  \State Calculate $\mathcal{L}_{state}$ and $\mathcal{L}_{action}$
  \State Calculate $\nabla g_{w_{t_{DM}}}$ from $\mathcal{L}_{total}$ and update $w_{t_{DM}}$
\EndFor
\State $w_{T_{DM}}$ will be the trained Decision Maker
\end{algorithmic}
\end{algorithm}

%% file: 04_data.tex
\section{Sepsis Data Description} \label{data}

The MIMIC-III database \cite{b29} is a comprehensive collection of de-identified clinical data of patients admitted to the Beth Israel Deaconess Medical Center in Boston, accessible to researchers globally under a data use agreement. It spans over a decade and contains data for 53,423 adult hospital admissions and 7870 neonate admissions, covering 38,597 distinct adult patients with a median age of 65.8 years and in-hospital mortality of 11.5\%. The database offers detailed information about patients' demographics, vital signs, laboratory test results, treatments, and outcomes, as well as free-text notes recorded by healthcare providers. The median length of an ICU stay is 2.1 days and a hospital stay is 6.9 days, with an average of 4,579 charted observations and 380 laboratory measurements available per hospital admission. The MIMIC-III database is widely used for research in critical care and offers a unique opportunity for researchers to study sepsis and other critical illnesses using a large, diverse, and rich dataset.

Sepsis MIMIC-III data is a subset of the MIMIC-III database, specifically focused on the patients who were diagnosed with sepsis during their ICU stay. It can be used to train models for sepsis detection and treatment, as well as for researching sepsis and its associated outcomes. 
The Sepsis MIMIC-III dataset comprises 19,614 treatment trajectories specific to sepsis patients. In order to design our experiment, we randomly allocate 30\% of this data for testing purposes, while the remaining 70\% is utilized for training. The following table presents a brief summary of this dataset.
Here, positive trajectories correspond to patient survival instances, while negative ones denote cases of patient fatality. The mortality rates for both the training and testing sets approximate 9.5\%.

\begin{table}[htbp]
\centering
\caption{Overview of the Sepsis Dataset}
\label{}
\begin{tabular}{|c||c|c|c|}
\hline
Type & Positive & Negative & Total \\
\hline\hline
Train & 12411 & 1319 & 13730\\
\hline
Test & 5321 & 563 & 5884\\
\hline
Total & 17732 & 1882 & 19614\\
\hline
\end{tabular}
\label{tab:data}
\end{table}

The Sepsis MIMIC-III dataset typically consists of several types of patient data, including demographic information, vital signs, laboratory test results, treatments, and outcomes. The input and output features that we are considering is listed as follows. \\
The data preprocessing phase comprised selecting pertinent features, discretizing them, and transforming them into a 4-hour window. Key physiological parameters, encompassing demographics, laboratory values, vital signs, and intake/output events, were systematically gathered for each patient. To facilitate analysis, the data were aggregated into 4-hour windows, with the mean or sum recorded—depending on the nature of the data—when multiple data points were available within a given window. Notably, we have used the same data preprocessing steps as detailed in \cite{raghu2017b}.

\textbf{Input features:}

\begin{itemize}
	\itemsep0em
	\item Demographic information: patient age, sex, weight, height.
	\item Vital signs: heart rate, blood pressure, temperature, oxygen saturation.
	\item Laboratory test results: white blood cell count, lactate level, blood glucose, creatinine.
	\item Treatments: medications administered, fluid boluses, mechanical ventilation.
	\item Length of stay: the number of days the patient spent in the ICU.
	\item Organ dysfunction: whether or not the patient experienced organ dysfunction during their ICU stay.    
\end{itemize}

\textbf{Output features:}

\begin{itemize}
	\itemsep0em
	\item Diagnosis: whether or not the patient was diagnosed with sepsis.
	\item Mortality: whether or not the patient died during their ICU stay.
\end{itemize}

% The dataset used in this study comprises 5321 positive trajectories, indicating cases where patients survived at the end, and 563 negative trajectories, representing cases where patients did not survive. Overall, the dataset exhibits a mortality rate of 9.5\% among all patients.

% comment: check and complete these features

% \dipesh{==========}\\

% The input features for our model include demographic information, vital signs, laboratory test results, treatments, length of stay, and organ dysfunction which makes our state space. The output features consist of the patient's recommended actions for trajectory management(action space), mortality (reward)

% the detailed preprocessing steps were performed as mentioned in \cite{b2}. The analysis in this paper was done on this preprocessed data.
% \dipesh{Might need to add in brief how \citep{b2} is handling the preprocessing for the completeness}

%% file: 05_evaluation.tex
\section{Experimental Results}
\label{sec:exp_results}

\begin{table}[htbp]
\caption{Hyperparameters for \name}
\centering
\begin{tabular}{|c | c|} 
 \hline
 Hyperparameter & Value  \\ 
 \hline\hline
 Number of layers & 3  \\ 
 \hline
 Number of attention heads & 1 \\
 \hline
 Embedding dimension & 128 \\
 \hline
 Batch size & 64  \\
 \hline
 Context Length & 3  \\
 \hline
 Nonlinearity & Relu  \\
 \hline
 Dropout rate & 0.1\\
 \hline
 Learning rate & $1e-4$\\
 \hline
Weight decay & $1e-4$\\
 \hline
 Warm-up steps & 10000\\
 \hline
$\alpha$: Weight of action prediction loss  & 1\\
 \hline
$\beta$: Weight of state prediction loss & 0.1\\
 \hline
$\gamma$: Weight of survival loss & 1\\
\hline

\end{tabular}
\label{param_expt}
\end{table}

In Table \ref{param_expt}, we present the hyperparameters used for \name, with the exception of those pertaining to the mortality classifier, which are outlined in Table \ref{cls_para}. The \tname model consists of three layers and utilizes a single attention head with an embedding dimension of 128. During training, we use a batch size of 64 and a context length of 3. The model employs the ReLU activation function as the non-linearity and includes a dropout layer with a rate of 0.1 to prevent overfitting. To avoid overfitting further, we set the learning rate to 1e-4 and apply a weight decay of 1e-4. The training process incorporates a warm-up phase of 10,000 steps, where the learning rate gradually increases from 0 to the specified value. The parameters $\alpha, \beta, \gamma$ control the weight of each term in the overall loss (i.e., Equation \eqref{eq:loss_total}). The hyperparameters, including the learning rate, number of epochs, and layers, underwent fine-tuning through an iterative process within a standard $k$-fold cross-validation setup. This procedure involved adjusting these settings and evaluating the model's performance on a validation dataset to identify configurations leading to the highest validation accuracy.

Assessing off-policy models presents a significant challenge as it is difficult to estimate the impact of executing a learned policy on patient mortality without a simulator. To address this challenge, we utilize a pretrained mortality classifier known for its high accuracy in predicting survival states. Our primary evaluation metric is the mortality rate, which indicates the proportion of patients experiencing negative outcomes as a result of the model's recommended treatments. It is important to note that the mortality rate considers the patient's death in any state along the trajectory.

In terms of baselines, we compare our \name model to two established approaches: the state-of-the-art offline reinforcement learning algorithm, Decision Transformer (DT) \cite{b1}, and the widely used imitation learning algorithm, Behavioral Cloning (BC) \cite{b15}. It is worth noting that other imitation learning algorithms like GAIL \cite{ho2016generative} and AIRL \cite{fulearning} require a simulator, making them unsuitable as baselines in our specific scenario.
Our code is publicly available at \url{https://github.com/Dipeshtamboli/PosNegDM-Reinforced-Sequential-Decision-Making-for-Sepsis-Treatment}.

\if 0
\begin{figure}
\begin{minipage}[c]{0.485\linewidth}
    \centering
    \includegraphics[width=\textwidth]{figs/state_min.png} 
    \label{fig:abl_state}
\end{minipage}
\begin{minipage}[c]{0.495\linewidth}
    \centering
    \includegraphics[width=\textwidth]{figs/max_loss_beta.png}
    \label{fig:max_loss_beta}
\end{minipage}
% \vspace{-10pt}
% \setlength{\belowcaptionskip}
\caption{ [WRONG FIG - NEED TO CHANGE THIS FIGURE]  (a) The plot shows the impact of different values of the hyperparameter $\lambda$ on the action prediction accuracy of \name model for positive and negative trajectories. The model was trained with different values of $\lambda$, which controls the contribution of state distance minimization in addition to the cross-entropy loss during training.
(b) The plot displays the action prediction accuracy of \name model trained with varying values of the $\beta$ hyperparameter, which controls the ratio between positive and negative trajectory cross-entropy loss during training. The findings highlight the trade-off between accurately predicting positive and negative outcomes when weighing the positive and negative loss differently, underscoring the importance of hyperparameter tuning during model training.
}
\label{fig:delay}
\end{figure}
\fi

\subsection{Achieving Low Mortality  with \name}
\label{sec:ach_low_mort}

\begin{table}[htbp]
\centering
\caption{Performance Comparison of Algorithms}
\label{performance_table}
\resizebox{\columnwidth}{!}{%
\begin{tabular}{ccccc}
\hline
\multirow{2}{*}{Algorithm} & \multirow{2}{*}{\makecell{Action Prediction Accuracy \\ on Positive Test Data\ \%}} &
  \multicolumn{3}{c}{Mortality \%} \\
  \cmidrule(lr){3-5} 
  & &
  \multicolumn{1}{c}{Positive Data} &
  \multicolumn{1}{c}{Negative Data} &
  Total \\ \hline
  \name & 94.6 & \multicolumn{1}{c}{\textbf{2.5}} & \multicolumn{1}{c}{\textbf{3.6}} & \multicolumn{1}{c}{\textbf{2.61}} \\\hline
Decision Transformer & 94.3 & \multicolumn{1}{c}{68.2} & \multicolumn{1}{c}{51.5}& \multicolumn{1}{c}{66.6}\\ \hline
Behavioral Cloning & \textbf{95.1} & \multicolumn{1}{c}{57.5} & \multicolumn{1}{c}{46.7} & \multicolumn{1}{c}{56.5}\\
\hline
\end{tabular}
}
\end{table}

Table \ref{performance_table} showcases the performance of our proposed algorithm, \name ($\alpha = 1, \beta = 0.1, \gamma = 1$) benchmarked against the Decision Transformer \citep{b1} and a Behavioral Cloning model \citep{b16}. The evaluation was conducted using test data from the Sepsis MIMIC-III dataset. The table includes several metrics: (i) action prediction accuracy for surviving patients in the test data (positive data), (ii) mortality rate among patients who were alive (positive data), (iii) mortality rate among patients who did not survive (negative data), and (iv) overall mortality rate. In order to obtain this metrics, we consider the last 10 time-steps of each trajectory. To assess action prediction on positive test trajectories, we compare the actions taken at each time-step in our approach with those in the expert data. Regarding mortality assessment, we input states from a test trajectory and observe whether the next predicted state indicates that the patient remains alive or not. The evaluation employs a trained mortality classifier in inference mode to make this determination. It is essential to clarify that the mortality classifier is trained independently on the training set and does not confer an advantage to the \name framework. The same mortality classifier is used consistently to assess the performances of Decision Transformer (DT) and Behavioral Cloning (BC).
This process is repeated for all states in the trajectory. If at any point, the patient is predicted not to be alive, it is classified as a negative mortality case. In contrast, if the patient is predicted to be alive throughout the trajectory, it is considered a positive mortality case. This mortality is called Step-by-step method, we will also consider a complete trajectory method in the next subsection. 

The results demonstrate a high action prediction accuracy of 94.6\% for the proposed algorithm, which is comparable to the baselines. However, it is important to note that the proposed algorithm significantly outperforms the baselines in terms of mortality rate. Simply achieving or mimicking actions is insufficient to prevent mortality, as any incorrect action within the trajectory can result in a patient not surviving.

The proposed algorithm, with its careful treatment to avoid mortality, achieves a remarkably low mortality rate of 2.61\%. In contrast, the Behavioral Cloning model had a mortality rate of 56.5\%, and the Decision Transformer model had a mortality rate of 66.6\%. It's significant to note that the figure 3.6\% suggests that with the application of our model's medical decisions, 96.4\% of patients, who otherwise would not have survived under the physicians' treatments, ultimately survived. This represents a substantial improvement. 
By mandating the prediction of state outputs and incorporating an adversarial mortality classifier, the model is empowered to enhance its mortality rate predictions. The feedback reinforcer, represented by the mortality classifier, plays a crucial role in offering additional guidance to the model, facilitating learning from errors. In totality, the amalgamation of these distinctive features and mechanisms is the driving force behind the superior performance of the \name framework when compared to existing machine learning-based algorithms.
These findings indicate that our proposed method is better suited for avoiding negative outcomes, such as mortality, which is a critical factor in clinical decision-making.

Therefore, the \name model holds the potential to significantly enhance the accuracy and reliability of clinical decision-making, ultimately improving patient outcomes in sepsis treatment.

\input{table_action}

\input{table_state}

\input{table_adv}

\subsection{Ablation studies}
\label{subsec:ablation_studies}

In this section, we conduct ablation studies to evaluate the impact of the three objective terms in the overall loss function (Equation \eqref{eq:loss_total}) on system performance, as delineated in Tables \ref{tab:abl_action} to \ref{tab:abl_adv}.

Here, we compare action prediction accuracy on the positive data and mortality.
Action prediction accuracy measures how well our model is able to mimic an expert's actions. Mortality gives us an idea of how well the model is able to drive patient states to alive states by taking proper actions. Given that we only have information from the trajectories available and no direct access to the model, we have calculated mortality in two different ways. In the Step-by-step method, we input states from a test trajectory and check if the next predicted state is an alive state or not. We do it for all the states in the trajectory.  In the Complete-trajectory state, we start with the initial test state and allow the model to generate the next ten actions and corresponding states. If any of the states results in a dead state, we consider that the patient has died in that trajectory. These mortality results are for both (positive and negative) test trajectories (i.e., total mortality).

Table \ref{tab:abl_action} showcases how the action prediction loss impacts the \name system's performance. Excluding $L_{action}$ (i.e., setting $\alpha=0$) results in a significant drop in action prediction accuracy on expert data, while increasing $\alpha$ enhances overall performance, as indicated by total mortality.

Table \ref{tab:able-state} reveals the effect of the subsequent state prediction loss on the \name system's performance. Without $L_{state}$ (i.e., setting $\beta=0$), subsequent state prediction and consequent mortality estimation from the classifier are unattainable. In addition, we can see that the total mortality increases as $\beta$ increases.

In Table \ref{tab:abl_adv}, we evaluate the impact of the survival loss on the performance of the \name system. We observe a significant increase in mortality when $L_{survival}$ is omitted, thereby underscoring its significance. Furthermore, we find that the performance with $\gamma = 0.8$ is not only comparable to that of $\gamma = 1.0$, but it actually results in better mortality results than the parameter setting we employed for Table \ref{performance_table}.

Overall, the results highlight the variability of mortality rates across different hyperparameters. 
The parameters \(\alpha\), \(\beta\), and \(\gamma\) provide control over \name's training objectives, allowing us to balance considerations such as prioritizing negative trajectory avoidance or achieving high action prediction accuracy. The impact of varying these parameters on the model's behavior and performance is demonstrated in Tables \ref{tab:abl_action}, \ref{tab:able-state}, and \ref{tab:abl_adv} for actions, states, and adversarial training, respectively.
Importantly, it should be noted that higher action prediction accuracy does not necessarily correlate with lower mortality rates. It is essential to explicitly study and consider mortality, which had not been  explored in the context of decision making prior to our work, to the best of our knowledge.

%% file: table_action.tex
\begin{table}[htbp]
\centering
\caption{Importance of the action prediction loss ($L_{action}$). Here, $\beta = 0.1$ and $\gamma = 1$.}
\label{tab:abl_action}
% \begin{tabularx}{\textwidth}{cccc}
\resizebox{.95 \columnwidth}{!}{%
\begin{tabular}{@{}c@{}c@{}c@{}c@{}}
\hline
\multirow{2}{*}{\makecell{Loss \\importance ($\alpha$)}} & \multirow{2}{*}{\makecell{Action Prediction Accuracy \\ on Positive Test Data\ \%}} &
  \multicolumn{2}{c}{Mortality \%} \\
  \cmidrule(lr){3-4} 
  & &
  \multicolumn{1}{c}{Step by Step} &
  \multicolumn{1}{c}{Complete Traj.} \\ \hline
  0.0 & 4.6 & \multicolumn{1}{c}{6.12} & \multicolumn{1}{c}{0.36} \\\hline
0.1 & 91.6 & \multicolumn{1}{c}{5.61}& \multicolumn{1}{c}{0.65}\\ \hline
0.3 & 94 & \multicolumn{1}{c}{6.07} & \multicolumn{1}{c}{0.63}\\ \hline
0.5 & 93.6 & \multicolumn{1}{c}{3.33} & \multicolumn{1}{c}{0.18} \\\hline
0.8 & 92.8 & \multicolumn{1}{c}{3.06}& \multicolumn{1}{c}{0.40}\\ \hline
1.0 & \textbf{94.6} & \multicolumn{1}{c}{\textbf{2.61}} & \multicolumn{1}{c}{\textbf{0.18}}\\ 
\hline
\end{tabular}
}
% \end{tabularx}
\end{table}

%% file: table_state.tex
\begin{table}[htbp]
\centering
\caption{Importance of subsequent state prediction loss ($L_{state}$). Here, $\alpha = 1$ and $\gamma = 1$.}
\label{tab:able-state}
\resizebox{.95\columnwidth}{!}{%
\begin{tabular}{@{}c@{}c@{}c@{}c@{}}
\hline
\multirow{2}{*}{\makecell{Loss \\importance ($\beta$)}} & \multirow{2}{*}{\makecell{Action Prediction Accuracy \\ on Positive Test Data\ \%}} &
  \multicolumn{2}{c}{Mortality \%} \\
  \cmidrule(lr){3-4} 
  & &
  \multicolumn{1}{c}{Step by Step} &
  \multicolumn{1}{c}{Complete Traj.} \\ \hline
  0.0 & 94.7 & \multicolumn{1}{c}{N$\backslash$A} & \multicolumn{1}{c}{N$\backslash$A} \\\hline
0.1 & 94.6 & \multicolumn{1}{c}{\textbf{2.61}}& \multicolumn{1}{c}{\textbf{0.18}}\\ \hline
0.3 & 92.9 & \multicolumn{1}{c}{3.57} & \multicolumn{1}{c}{0.36}\\ \hline
0.5 & 93.5 & \multicolumn{1}{c}{6.33} & \multicolumn{1}{c}{0.40} \\\hline
0.8 & 92.9 & \multicolumn{1}{c}{7.54}& \multicolumn{1}{c}{0.20}\\ \hline
1.0 & 92.8 & \multicolumn{1}{c}{10.02} & \multicolumn{1}{c}{0.67}\\ 
\hline
\end{tabular}
}
\end{table}

%% file: table_adv.tex
\begin{table}[htbp]
\centering
\caption{Importance of the survival loss ($L_{survival}$). Here, $\alpha = 1$, $\beta = 0.1$.}
\label{tab:abl_adv}
\resizebox{.95\columnwidth}{!}{%
\begin{tabular}{@{}c@{}c@{}c@{}c@{}}
\hline
\multirow{2}{*}{\makecell{Loss \\importance ($\gamma$)}} & \multirow{2}{*}{\makecell{Action Prediction Accuracy \\ on Positive Test Data\ \%}} &
  \multicolumn{2}{c}{Mortality \%} \\
  \cmidrule(lr){3-4} 
  & &
  \multicolumn{1}{c}{Step by Step} &
  \multicolumn{1}{c}{Complete Traj.} \\ \hline
  0.0 & \textbf{94.7} & \multicolumn{1}{c}{60.41} & \multicolumn{1}{c}{10.29} \\\hline
0.1 & 94.3 & \multicolumn{1}{c}{5.21}& \multicolumn{1}{c}{0.49}\\ \hline
0.3 & 93.9 & \multicolumn{1}{c}{3.95} & \multicolumn{1}{c}{0.45}\\ \hline
0.5 & 93.4 & \multicolumn{1}{c}{\textbf{2.32}} & \multicolumn{1}{c}{0.36} \\\hline
0.8 & 93.9 & \multicolumn{1}{c}{2.41}& \multicolumn{1}{c}{\textbf{0.18}}\\ \hline
1.0 & 94.6 & \multicolumn{1}{c}{2.61} & \multicolumn{1}{c}{\textbf{0.18}}\\ 
\hline
\end{tabular}
}
\end{table}

%% file: 06_conclusion.tex
\section{Conclusion}
\label{sec:conclusion}
Sepsis is a severe medical condition where rapid, precise interventions can significantly decrease mortality rates. Existing guidelines, however, fall short of providing real-time, individualized treatment decisions. Reinforcement Learning (RL) has shown promise in the medical field by modeling physicians' decision-making processes but relies on a simulator. To circumvent these challenges, we proposed the \name framework, an Offline RL-based solution. This innovative method assists physicians in devising effective treatment strategies for sepsis patients by learning from experts' decisions (e.g., MIMIC-III dataset). The architecture of \name includes a Transformer model and a feedback reinforcer, enabling the agent to learn from successful decision sequences while veering away from those leading to negative outcomes. Our model could replicate expert decision-making with 94.6\% accuracy and showed an improvement in survival rates, saving 97.39\% of patients compared to survival rates of 33.4\% (DT) and 43.5\% (BC). The \name framework hence emerges as a promising tool in assisting physicians in making informed, timely treatment decisions for sepsis patients, underscoring its potential in significantly improving patient outcomes.

Our work shows promise in enhancing sepsis treatment, with future considerations for deploying the model in clinical practice. Ethical and legal aspects, such as patient privacy and informed consent, require careful attention. The model may need  periodic adaptation to address real-world clinical complexities. We also acknowledge limitations, including dependence on training data quality, generalization challenges, and interpretability issues, as directions for future work.

%% file: 071_appendix.tex
 \appendices

 \if 0
 \setcounter{page}{1}
\fi 
%\section{Appendix}
\onecolumn
%{\bf Appendices for ``Reinforced Sequential Decision-Making for Sepsis Treatment: The \name Framework with Mortality Classifier and Transformer"}
\section{Enlarged Figure for Figure \ref{fig:intro}}

\begin{figure*}[ht]
\centerline{\includegraphics[scale=0.2]{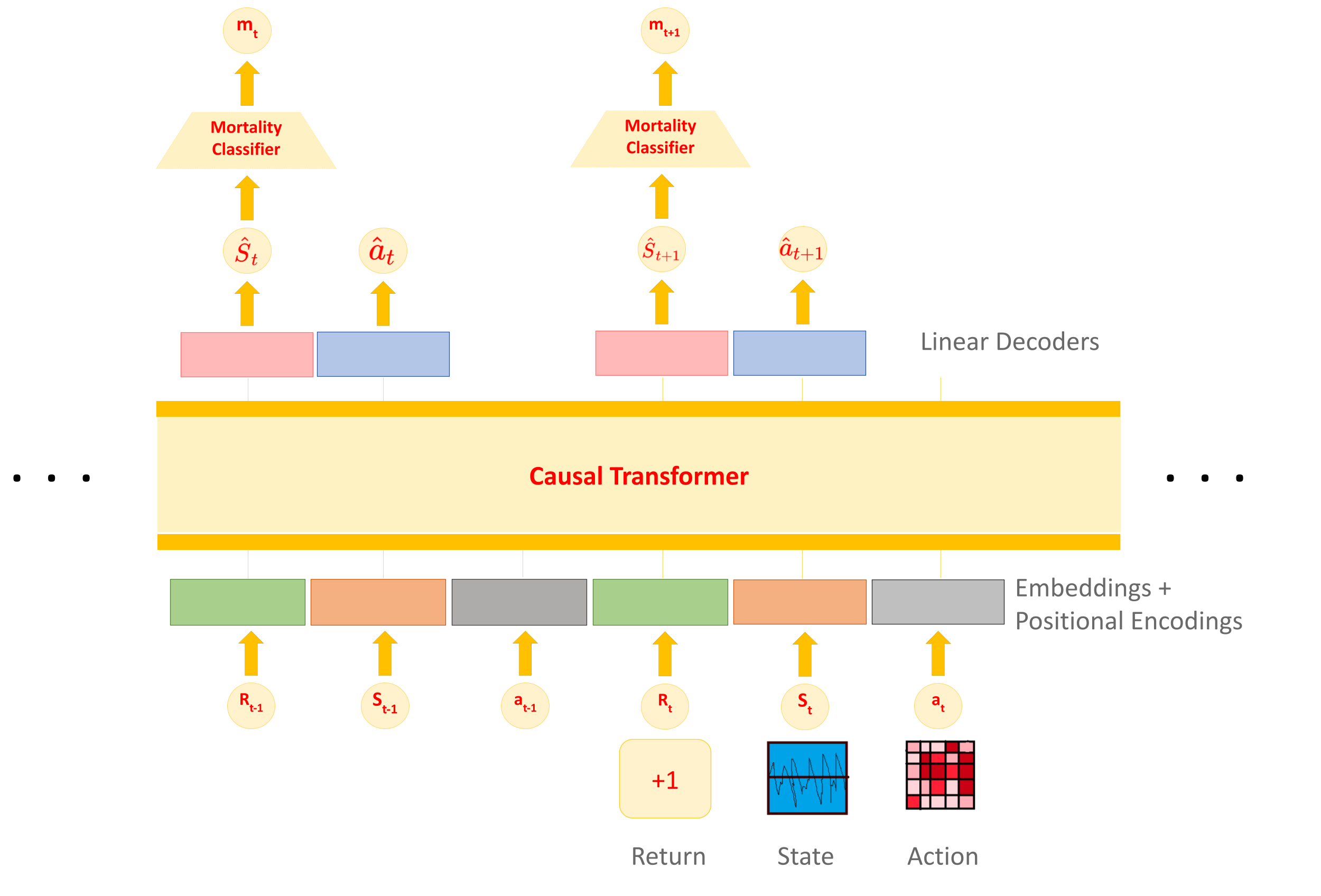}}
	\caption{The \tname decision maker takes in states, actions, and returns as input, which are first embedded into linear representations that are specific to each modality. The positional episodic timestep encoding is added to the input to help the model understand the order of events. The tokens are then fed into the GPT architecture, which uses a self-attention mechanism to predict actions and next states. The causal mask ensures that the model can only attend to previous tokens, preserving the causality of the system. The predicted states are subsequently input into the trained Mortality Classifier to assess whether the implemented action guides the patient towards a deceased state. The mortality prediction \(m_t\) is employed to influence the \tname, compelling it to choose actions aligned with the mortality classifier's prediction of an alive state. This process integrates mortality considerations into the decision-making mechanism, emphasizing the importance of actions that contribute to favorable patient outcomes.}
	\label{fig:intro2}
\end{figure*}

The enlarged version of Figure \ref{fig:intro} is depicted in Figure \ref{fig:intro2}. 

\section{Additional experiments}
\label{sec:add_exps}
\input{table_sensitivity}

This section includes the sensitivity experiments and additional 2D histograms that visualize the aggregated actions recommended by the physician (ground truth), \name, and Behavioral Cloning (BC). These histograms provide a qualitative analysis of the actions recommended by each method and highlight the differences in decision-making strategies.

In Table \ref{tab:sensitivity}, we present the results of a sensitivity experiment conducted to assess the stability of our model across different random seeds. The experiment involves five runs with varied random seeds, maintaining fixed hyperparameters ($\alpha = 1$, $\beta = 0.1$, $\gamma = 1$). The table reports the action prediction accuracy on positive test data and mortality rates using both step-by-step and complete trajectory methods. The mean ($\mu$) and standard deviation ($\sigma$) across the five runs are provided, demonstrating consistent performance with an average action prediction accuracy of 94.36\%, a mortality rate of 2.29\% (step-by-step), and 0.11\% (complete trajectory). These results underscore the robustness of our model, indicating minimal variability in outcomes across different random seeds.

\section{Understanding the Action Space}
\begin{figure*}[t]
\centering
\subfigure[Ground Truth (Positive)]{
\label{fig:2(a)} 
\includegraphics[width=2.5in, height=2.0in]{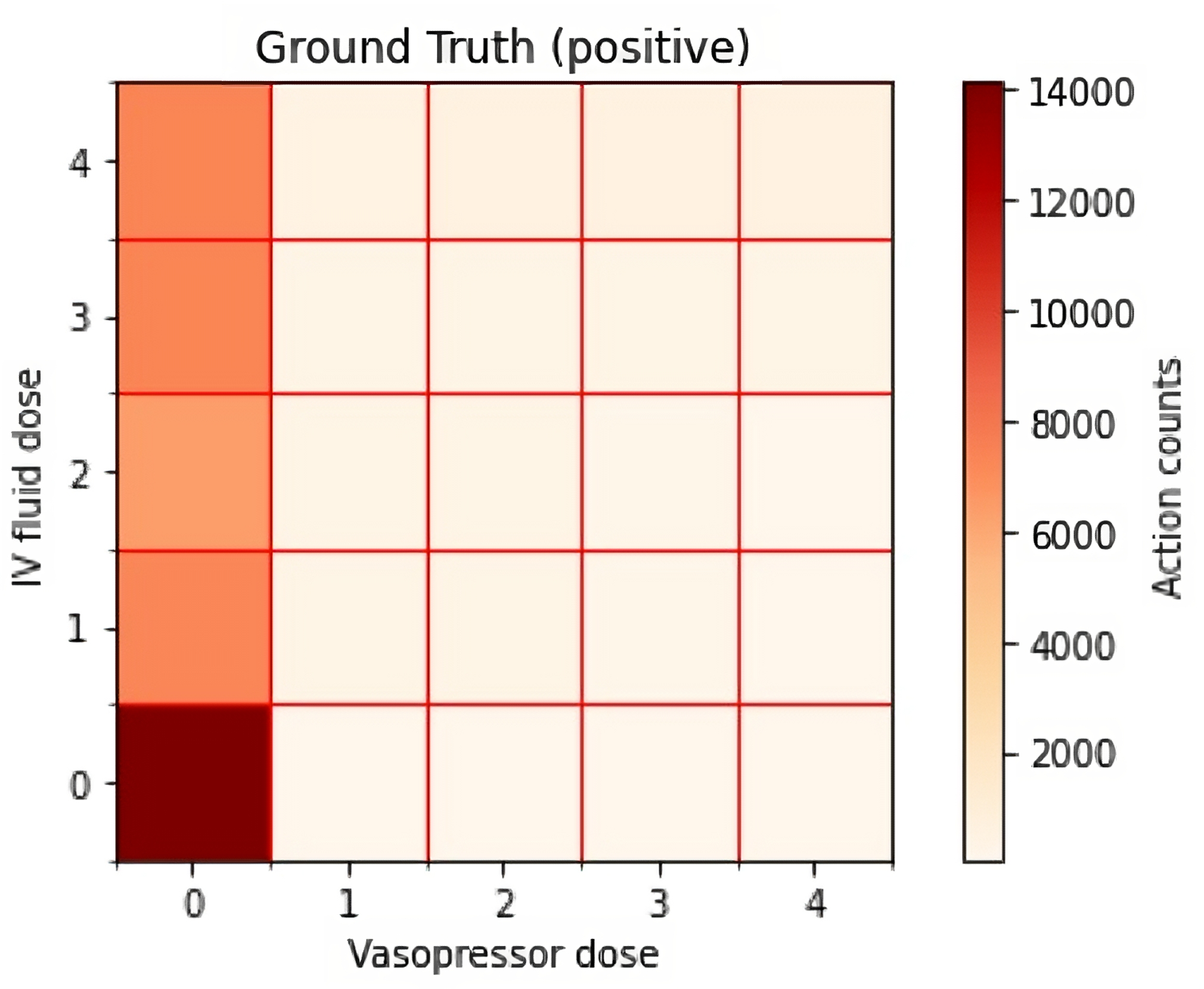}}
\subfigure[Ground Truth (Negative)]{
\label{fig:2(b)} 
\includegraphics[width=2.5in, height=2.0in]{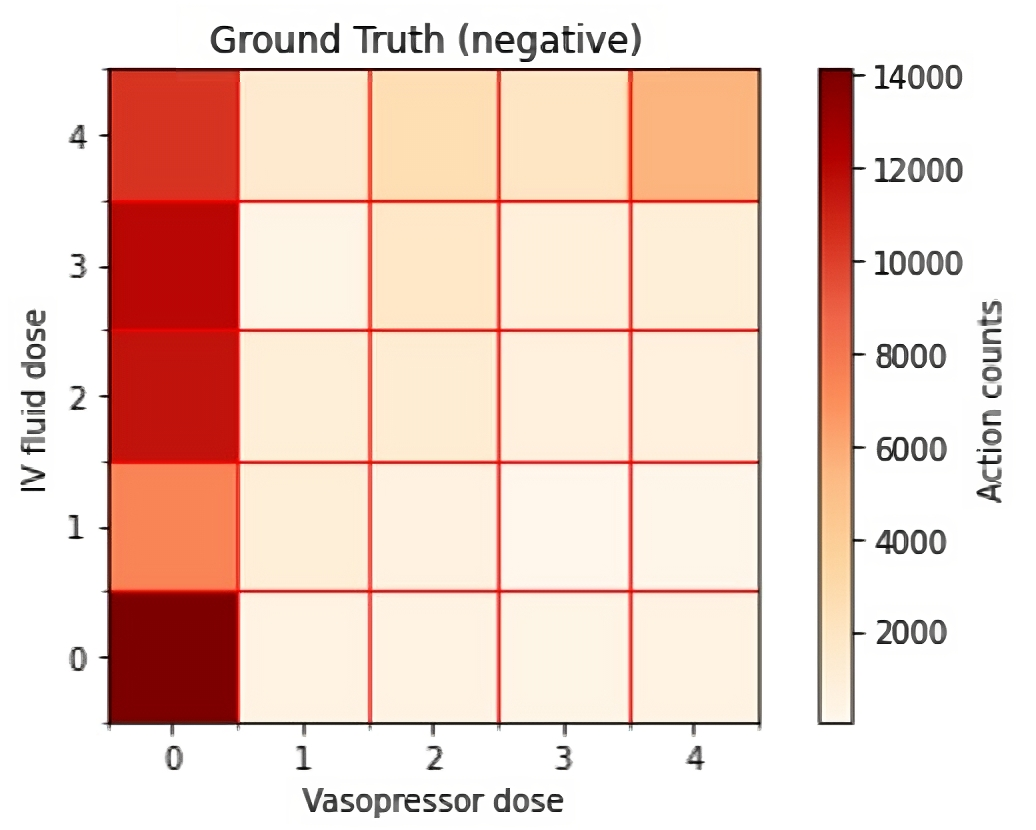}}
\subfigure[\name (Positive)]{
\label{fig:2(c)} 
\includegraphics[width=2.5in, height=2.0in]{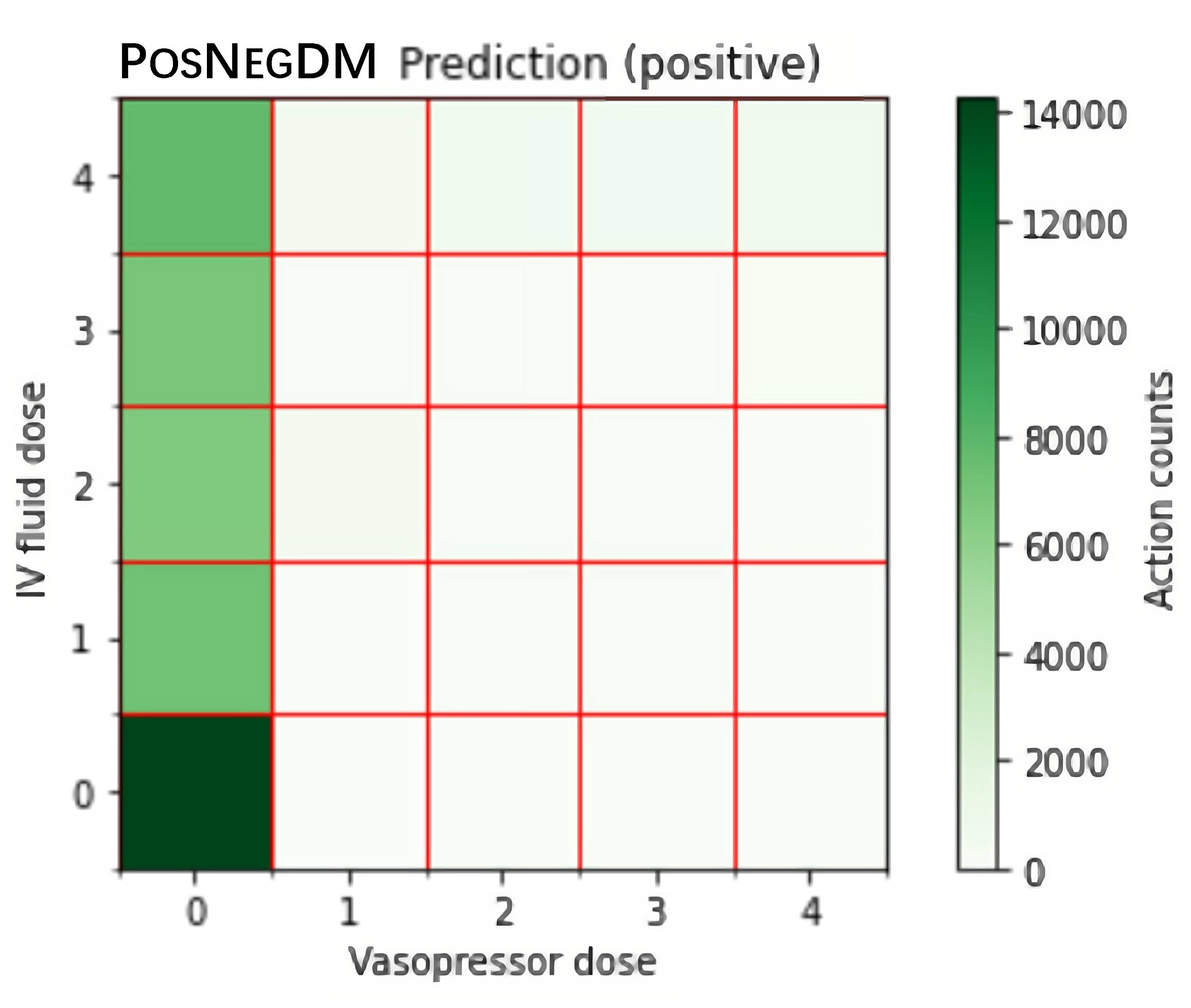}}
\subfigure[\name (Negative)]{
\label{fig:2(d)} 
\includegraphics[width=2.5in, height=2.0in]{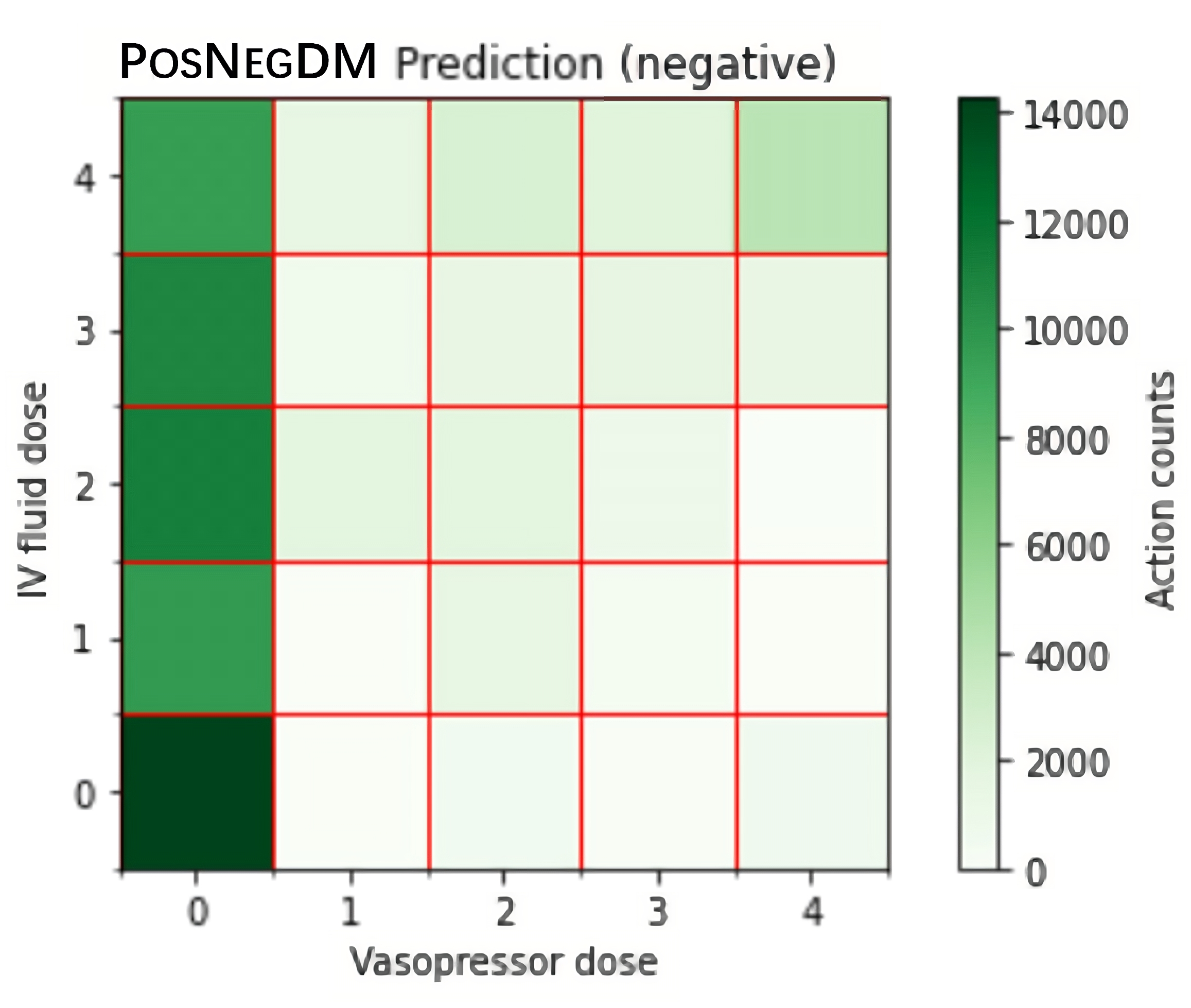}}
\subfigure[BC (Positive)]{
\label{fig:2(e)} 
\includegraphics[width=2.5in, height=2.0in]{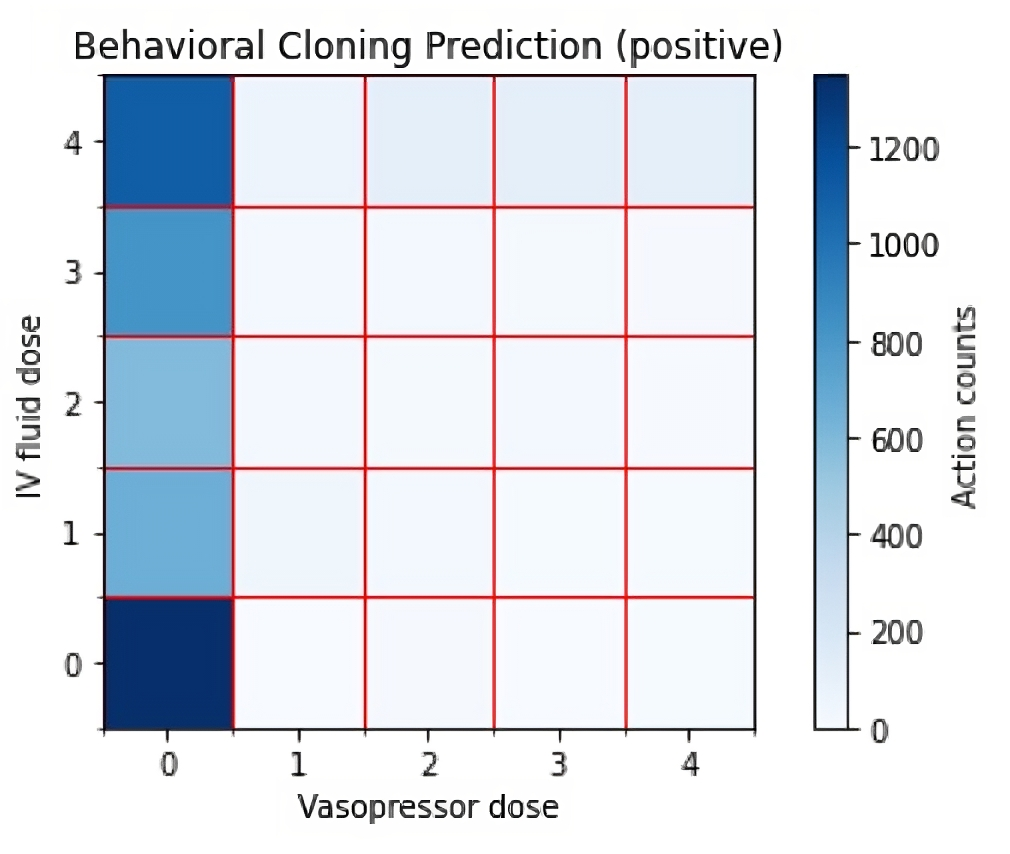}}
\subfigure[BC (Negative)]{
\label{fig:2(f)} 
\includegraphics[width=2.5in, height=2.0in]{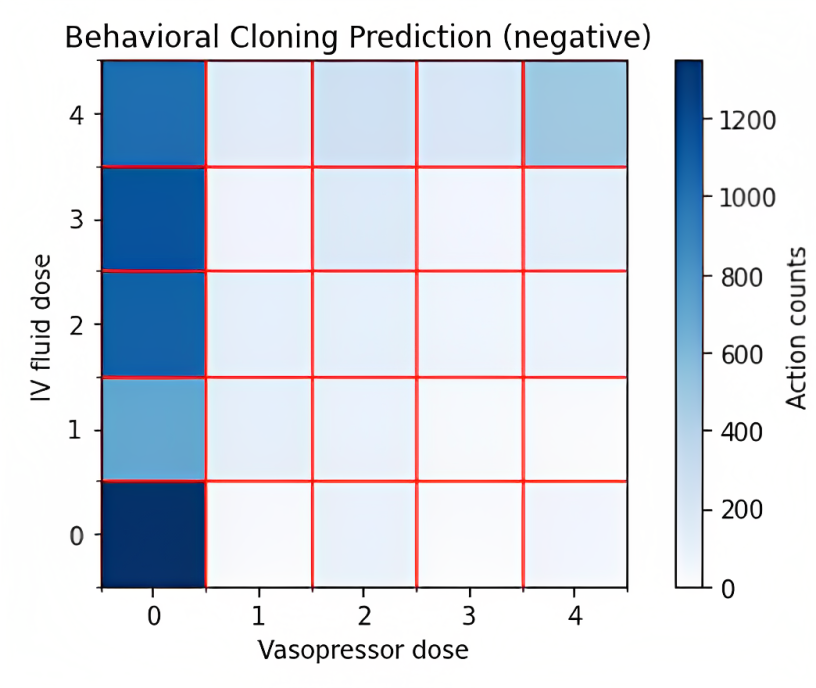}}
\caption{The three rows in the visualization represent the policies as provided by physicians, \name, and Behavioral Cloning (BC) respectively, each applied to both positive and negative test data. The axis labels correspond to the discretized action space, where '0' signifies no drug administration, and '4' indicates the maximum dosage of a particular drug. Each grid cell represents a specific action, with its color indicating the frequency of its occurrence.}
\label{fig:2} 
\end{figure*}

Figure \ref{fig:2} presents 2D histograms that visualize the aggregated actions recommended by the physician (Ground Truth), \name, and Behavioral Cloning (BC). These histograms are analogous to those used for qualitative analysis in \cite{raghu2017b}. The horizontal and vertical bins correspond to the prescribed dosages of Vasopressor and IV fluid, respectively, based on the given policy. An action value of 0 signifies that no drugs are administered to the patient, while an increasing action value corresponds to a higher dosage. These non-zero drug dosages are represented in quartiles. Each grid cell denotes a specific action, with the color reflecting its frequency of occurrence. Notably, the \name model appears to mirror the ground truth more accurately for positive cases than the BC model does, a distinction especially noticeable in the action tuple (0,4) for the positive data. The results from Decision Transformer is not included, since it deviates from the ground truth to an even greater extent than Behavioral Cloning.

%% file: table_sensitivity.tex
\begin{table}[htbp]
\centering
\caption{Sensitivity experiment to check stability of results with 5 runs with different random seeds for $\alpha = 1$ Here, $\beta = 0.1$ and $\gamma = 1$.}
\label{tab:sensitivity}
\begin{tabular}{@{}cc@{}c@{}c@{}}
\hline
\multirow{2}{*}{\makecell{Random\\ Seed}} & \multirow{2}{*}{\makecell{Action Prediction Accuracy \\ on Positive Test Data\ \%}} &
  \multicolumn{2}{c}{Mortality \%} \\
  \cmidrule(lr){3-4} 
  & &
  \multicolumn{1}{c}{Step by Step} &
  \multicolumn{1}{c}{Complete Traj.} \\ \hline
1 & 94.6 & \multicolumn{1}{c}{2.61}& \multicolumn{1}{c}{0.18}\\ \hline
2 & 93.6 & \multicolumn{1}{c}{1.89} & \multicolumn{1}{c}{0.18}\\ \hline
3 & 94.1 & \multicolumn{1}{c}{2.61} & \multicolumn{1}{c}{0.0} \\\hline
4 & 94.9 & \multicolumn{1}{c}{2.04}& \multicolumn{1}{c}{0.18}\\ \hline
5 & 94.6 & \multicolumn{1}{c}{2.29} & \multicolumn{1}{c}{0.0}\\ \hline
 \bm{$\mu \pm \sigma$}  & \textbf{94.36 $\pm$ 0.51} & \multicolumn{1}{c}{\textbf{2.29$\pm$0.33}} & \multicolumn{1}{c}{\textbf{0.11$\pm$0.1}}\\ 
\hline
\end{tabular}
\end{table}